\documentclass[runningheads]{llncs}

\usepackage{graphicx}
\usepackage{subcaption}
\usepackage{amssymb}
\usepackage{amsmath}
\usepackage{siunitx}
\usepackage{xcolor}
\usepackage[belowskip=0pt,aboveskip=0pt]{caption}
\usepackage[belowskip=0pt,aboveskip=0pt]{subcaption}
\usepackage{mathtools}
\DeclarePairedDelimiterX\set[1]\lbrace\rbrace{#1}

\begin{document}

\title{Learning Personal Representations from fMRI by Predicting Neurofeedback Performance}

\titlerunning{fMRI Personal Representation}

\author{Jhonathan Osin,\inst{1} Lior Wolf\inst{1}\\
Guy Gurevitch,\inst{2, 3} Jackob Nimrod Keynan,\inst{2, 3} Tom Fruchtman-Steinbok,\inst{2} \\ Ayelet Or-Borichev,\inst{2, 4} Shira Reznik Balter\inst{2} and Talma Hendler\inst{2, 3,4, 5}}

\authorrunning{J. Osin, L. Wolf et al.}

\institute{School of Computer Science, Tel Aviv University \and
Sagol Brain Institue, Tel-Aviv Sourasky Medical Center \and
School of Psychological Sciences, Tel Aviv University \and
Sackler Faculty of Medicine, Tel Aviv University \and
Sagol School of Neuroscience, Tel Aviv University
}

\maketitle

\begin{abstract}
We present a deep neural network method for learning a personal representation for individuals that are performing a self neuromodulation task, guided by functional MRI (fMRI).

This neurofeedback task (watch vs. regulate) provides the subjects with a continuous feedback contingent on down regulation of their Amygdala signal and the learning algorithm focuses on this region's time-course of activity. The representation is learned by a self-supervised recurrent neural network, that  predicts the Amygdala activity in the next fMRI frame given recent fMRI frames and is conditioned on the learned individual representation. 

It is shown that the individuals' representation improves the next-frame prediction considerably. Moreover, this personal representation, learned solely from fMRI images, yields good performance in linear prediction of psychiatric traits, which is better than performing such a prediction based on clinical data and personality tests. Our code is attached as supplementary and the data would be shared subject to ethical approvals.
\end{abstract}

\section{Introduction}
In this work, we propose to employ self-supervision in order to learn an individual, per-subject representation from fMRI-based neurofeedback sessions. 
Neurofeedback (NF) is a Brain Computer Interface approach for non-invasive self-neuromodulation via reinforcement learning. NF has been widely used in the last decade in research and clinical settings for training people how to alter their own brain functionality; activity or connectivity.  

This representation is shown to be highly predictive of multiple psychiatric condition in three different datasets: (i) individuals suffering from PTSD, (ii) individuals suffering from fibromyalgia and (iii) a control dataset of healthy individuals. 

The self-supervised method predicts the activity of the Amygdala at the next fMRI frame based on the previous frames, conditioned on the individual representation. For this purpose, we employ a variant of the LSTM algorithm, in which the personal embedding is used to condition all four LSTM gates.  

The learned personal embedding is a static vector, which encodes information about the individual that is meaningful in predicting the future state of the Amygdala. Remarkably, this vector is more predictive of the subject's psychiatric traits, age and previous experience in neurofeedback, than the individual's data, when predicting one trait from all other traits.

The comparison is done using linear classifiers, in order to verify that the relevant information is encoded in a relatively explicit way and to alleviate the risk of over-fitting by attempting multiple hyper-parameters.

\begin{figure}[t]
    \centering
    \includegraphics[width=0.9\textwidth]{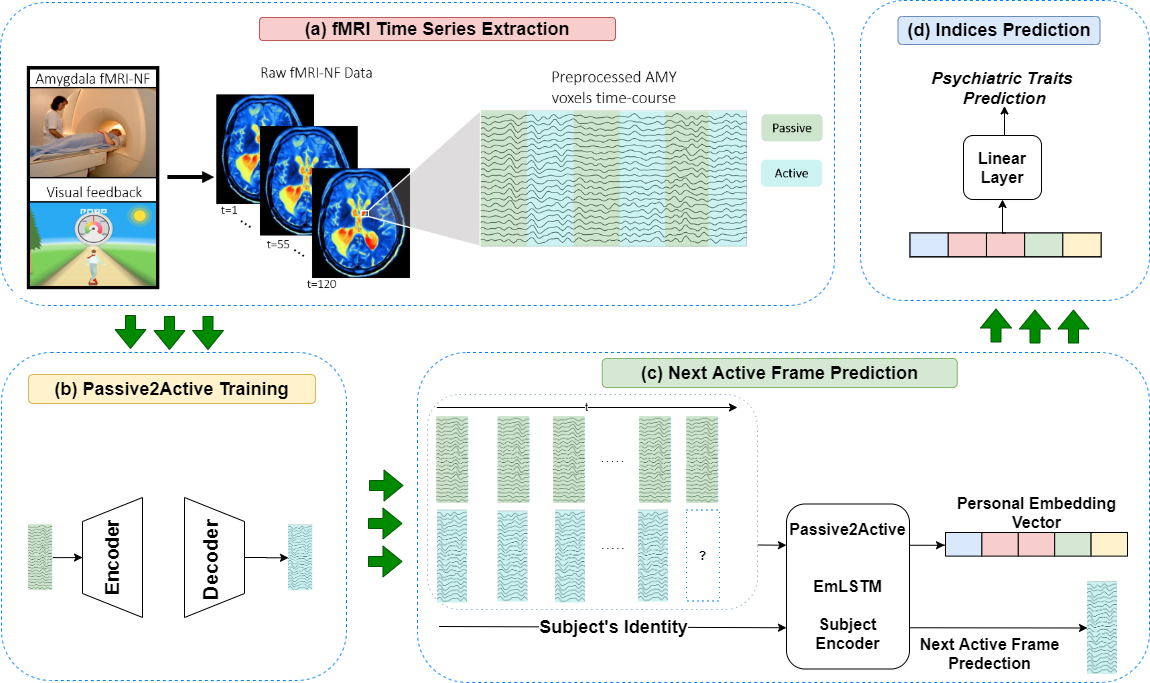}
    \caption{Holistic overview of our method. (a) The time series are extracted from raw data. (b) The Passive2Active Translator is trained, without knowledge of subject's identity. (c) subject-conditioned LSTM and a look-up-table (LUT) are trained via the next active frame prediction task. (d) A linear classifier is trained to predict psychiatric and demographic traits based on learned per-subject representations.}
    \label{fig:algo: overview}
\end{figure}

A complete overview of our approach is illustrated in Fig.~\ref{fig:algo: overview}. The first step after extracting fMRI frames, is to learn an auto-encoder which maps between the two stages of the neurofeedback session; one in which the subject is viewing passively, and one in which the subject is requested to control their Amygdala activity. A new variant of LSTM is then trained to predict the activity in the next fMRI frame. Finally, the representation learned as part of the LSTM is employed for predicting psychiatric traits.

\section{Related Work}

Functional magnetic resonance imaging (fMRI), as a non-invasive imaging technique, has been extensively applied to study psychiatric disorders~\cite{calhoun2014chronnectome,oksuz2019magnetic}. Deep learning methods have been applied to this field, mostly focusing on binary classification of subjects suffering from a specific psychopathology versus healthy subjects under resting state~\cite{dvornek2017identifying,yan2019discriminating}, i.e., using fully supervised learning. Our work is focused on self-supervised learning while participants suffering from various psychopathologies and healthy controls are performing a neurofeedback training task.

During an NF session, the trainee is given the task of regulating their brain state in a target region using some mental strategy. Brain activity modulation in the determined direction (up or down), resulting in contingent change in a rewarding perceived interface, thus reflecting level of task success ~\cite{lubianiker2019process}. Recent technological advancements in online real-time data analysis have made fMRI a popular tool for employing NF in research settings, providing brain targeting with high spatial resolution ~\cite{sulzer2013real}. However, due to the relatively high cost of using this imaging modality and the burden of multiple sessions, fMRI-based NF could benefit from on-line personalization of the procedure, which would make the training more efficient and advance this tool into translational and clinical trials.

\section{Problem Formulation}
A real-time fMRI-NF task targeting down-regulation of the right Amygdala region was given, using an interactive game-like feedback interface where the subjects control the speed of a skateboard rider. Similar to other studies in the field, local fMRI activation changes were measured using a two-phase NF paradigm repeated over several runs~\cite{paret2019current}, where each training run was comprised of a passive and an active phase. During the passive phase, a skateboard rider and a speedometer were displayed on the screen and Amygdala activation was passively measured. Subjects were instructed to passively view the skateboard, which was moving at a constant speed. During the following active phase, the speed of the rider and the speedometer represented the on-going Amygdala signal change compared to the passive phase, which was calculated on-line and updated continuously every three seconds.

During this phase, subjects were instructed to decrease the speed of the skateboard as much as possible, by practicing mental strategies of their choosing. Down-regulation of the Amygdala, which reflects a more relaxed state, led to a lower skateboard speed and fewer objects on the screen~\cite{lubianiker2019process}. Instructions given were not specific to the target brain functionality, in order to allow individuals to efficiently adopt different strategies~\cite{marxen2016amygdala}. At the end of each active phase, a bar indicating the average speed during the current run was presented for six seconds. Each subject performed $M=3$ runs of Passive/Active phases, where each passive phase lasted one minute. Each active phase lasted one minute (Healthy controls and PTSD patients) or two minutes (Fibromyalgia patients). 

    Both the passive and the active phases gave rise to $T=14$ temporal samples of the subject's Amygdala. We denote the spatial resolution of the relevant part of the fMRI images acquired every 3 seconds as $H \times W \times D$ and our dataset is, therefore, comprised of per-subject tensors in $\mathbb{R}^{2\times H\times W\times D\times T\times M}$. In our setting, $M=3, T=14$.
We used three datasets in our experiments: (i) \textbf{PTSD}- comprised of 53 subjects, $(H=6, W=5, D=6)$, (ii) \textbf{Fibromialgya}- comprised of 24 subjects, $(H=6, W=5, D=6)$, and (iii) \textbf{Healthy Control}- comprised of 87 subjects, $(H=10, W=8, D=10)$.

In addition to the fMRI sequences, we receive clinical information about each subject, which we denote as $y_n\in\mathbb{R}^3$, $n$ being the index of the patient, and is comprised of the following: (1) \textbf{Toronto Alexithymia Scale (TAS-20)} – a self-report questionnaire measuring difficulties in expressing and identifying emotions~\cite{bagby1994twenty}, (2) \textbf{State-Trait Anxiety Inventory (STAI)} – State anxiety was measured using a validated 20-item inventory~\cite{spielberger1983state}, and (3) \textbf{Clinician Administered PTSD Scale (CAPS-5) 1} – Patients underwent clinical assessment by a trained psychologist based on this widely-used scale for PTSD diagnosis~\cite{weathers2013clinician}. For the control patients, we also receive the following demographic information: (1) \textbf{Age} and (2) \textbf{Past Experience in Neuro-Feedback Tasks}, quantized to three levels: no experience, two previous sessions, or six previous sessions.

\noindent{\bf fMRI Data Acquisition and Pre-Processing\quad}
Structural and functional scans were performed in a 3.0T Siemens MRI system (MAGNETOM Prisma) using a 20-channel head coil. To allow high-resolution structural images, a T1-weighted three-dimensional (3D) sagittal MPRAGE pulse sequence (repetition time/echo time=1,860/2.74 ms, flip angle=\ang{8}, pixel size=$1\times 1$ mm, field of view=$256\times 256mm$) was used. Functional whole-brain scans were performed in an interleaved top-to-bottom order, using a T2*-weighted gradient echo planar imaging pulse sequence (repetition time/echo time=3,000/35ms, flip angle=\ang{90}, pixel size=1.56mm, field of view=$200\times 200 mm$, slice thickness=3 mm, 44 slices per volume). 

Preprocessing of the fMRI data was done with the MATLAB based CONN toolbox~\cite{whitfield2012conn} and included realignment of the functional volumes, motion correction using rigid-body transformations in six axes, normalization to MNI space and spatial smoothing with an isotropic 6-mm full width at half-maximum Gaussian kernel. The processed volumes were then run through de-noising and de-trending regression algorithms, followed by band-pass filtering in the range of of 0.008-0.09 Hz. Amygdala voxels were defined as a functional cluster centered at (x=21, y=-2, z=-24) and exported for further analysis as a 4-D matrix.

\section{Method}
Our neural network models were trained to predict the subjects' clinical and demographic information, given the raw fMRI sequences. This was done using two sub-networks, denoted as $f, \rho$ respectively: (1) Learning a personalized representation for each subject, and (2) predicting the subjects' clinical and demographic information, given  the representation learned by $f$.

\subsection{Overview}
The task performed by sub-network $f$ is predicting the current active frame, given all previous passive and active frames and the current passive frame. 

{\noindent{\bf Next Active Frame Prediction \quad}} For each subject $n$, active and passive frames from the $M$ sessions are concatenated (separately) to receive the entire passive and active fMRI \textbf{sequences}, denoted as $p^n, a^n \in \mathbb{R}^{(H, W, D, M\cdot T)}$, respectively. Thus $p^n[t], a^n[t]$ are the passive and active fMRI \textbf{frames} captured at time step $t$. 

Capital letters $P^n[t] = (p^n[0], p^n[1],\dots, p^n[t])$,and $A^n[t] = (a^n[0], a^n[1],\dots, a^n[t])$ are used to denote a sequence from its beginning, and until time step $t$. The model $f$ predicts the active-phase tensor at time point $t$ of subject $n$: 
\begin{equation}
\hat{a}^n[t] = f(A^n[t-1], P^n[t], n)
\label{eq: f implicit}
\end{equation}

\noindent{\bf Predicting Subjects' Traits\quad} A by-product of the training process of $f$, is a learned per-subject personal embedding {vector}, denoted as $e_n$. The prediction of $\rho$ for subject $n$ can be denoted as:
\begin{equation}
    \hat{y}_n = \rho(e_n) = G^{\top} e_n + b\, \label{eq:rho}
\end{equation}
where $G$ is a {matrix and $b$ is a vector}.

\subsection{Learning a Personalized Representation}
The $f$ model is composed of two deep networks. The first network, $\varphi$, maps, one by one, passive frames to the corresponding active frames. The second sub-network, $\psi$, is an LSTM that predicts the current active frame, given the output of the first sub-network, the previous active frame and the subject's identifier.

\subsubsection{Passive2Active translator}
This model, $\varphi$, is trained in a supervised way to map an input frame from the passive phase, $p^n[t]$, to its coupled frame from the active phase, $a^n[t]$, with the subject's identity unknown to the model. Training is done by minimizing the following loss function: 
\begin{equation}
    \mathcal{L}_{Recon} = \sum_n\sum_{t=0}^{M\cdot T}||a^n[t] - \varphi(p^n[t])||_2
    \label{eq: p2a loss}
\end{equation}

The architecture follows that of de-noising autoencoders, which are widely used in medical image analysis~\cite{gondara2016medical}. The network $\varphi$ is comprised of four linear layers with expanding output sizes, followed by four shrinking linear layers, separated by DropOut and ReLU activation functions, alternately.

\subsubsection{The conditioned LSTM network}
In the conventional LSTM~\cite{hochreiter1997long}, given an input $x_t$ = $\big(\varphi(p^n[t]), a^n[t-1]\big)$, the previous hidden state $h_{t-1}$, and the previous cell state $c_{t-1}$, the LSTM's outputs $c_t$ and $h_t$ are calculated using four learned gates, dubbed \textit{forget gate, input gate, update gate, output gate}. The information flow is as depicted in Eq.4--9. In the conditioned LSTM we employ, a representation is learned for every subject $n$, marked $e_n$. This user-embedding is used as a conditioning input to the LSTM sub-networks, by concatenating it to the input at every time step. In this setting, the information flow is as depicted by equations Eq.4*--7*, Eq.8--9.

\noindent\begin{tabular}{lcl}
\textsc{Vanilla Lstm Equations \hspace{1cm}} &\quad\quad &\textsc{Conditioned LSTM Equations} \\
$f_t^n = \sigma(W^f\cdot(x_t, h_{t-1}))$ \hspace{1.05cm} (4) & & $f_t^n = \sigma(W^f\cdot(x_t, h_{t-1}, e_n))$ \hspace{1.0cm}(4*)\\
$i_t^n = \sigma(W^i\cdot(x_t, h_{t-1}))$ \hspace{1.2cm} (5) & & $i_t^n = \sigma(W^i\cdot(x_t, h_{t-1}, e_n))$ \hspace{1.15cm}(5*)\\
$u_t^n = \tanh(W^u\cdot(x_t, h_{t-1}))$ \hspace{0.55cm} (6) & & $u_t^n = \tanh(W^u\cdot(x_t, h_{t-1}, e_n))$ \hspace{0.55cm}(6*)\\
$o_t^n = \sigma(W^o\cdot(x_t, h_{t-1}))$ \hspace{1.1cm} (7) & & $o_t^n = \sigma(W^o\cdot(x_t, h_{t-1}, e_n))$ \hspace{1.05cm}(7*)\\
\multicolumn{3}{c}{$c_t^n = c_{t-1}^n\odot f_t^n + u_t^n\odot i_t^n$ \hspace{2cm}(8)}\\
\multicolumn{3}{c}{$h_t^n = o_t^n \odot \tanh(c_t^n)$ 
\hspace{2.8cm}(9)}\\
\end{tabular}\\
\noindent Where $W^{(\cdot)}$ are learned weights for each one of the gates, $\sigma(\cdot)$ is the Sigmoid function, $\odot$ is the element-wise multiplication operator and $(\cdot, \cdot)$ is the concatenation operator.

Once the first sub-network, $\varphi$, is trained, it is frozen and we train, concurrently, the second sub-network, $\psi$, and the subjects' encoding vectors, which are stored in a look-up-table (LUT) $\chi$. With all sub-networks defined, we can write Eq.~\ref{eq: f implicit} more explicitly:
\begin{equation}
    \hat{a}^n[t] = f\big(A^n[t-1], P^n[t], n\big) =  \psi\big(A^n[t-1],\varphi(P^n[t]), \chi(n)\big)\, ,
    \label{eq: f explicit}
    \tag{10}
\end{equation}
where $\varphi$ is applied separately for every frame of $P^n[t]$.

\begin{figure}[t]
    \centering
    \includegraphics[width=\textwidth]{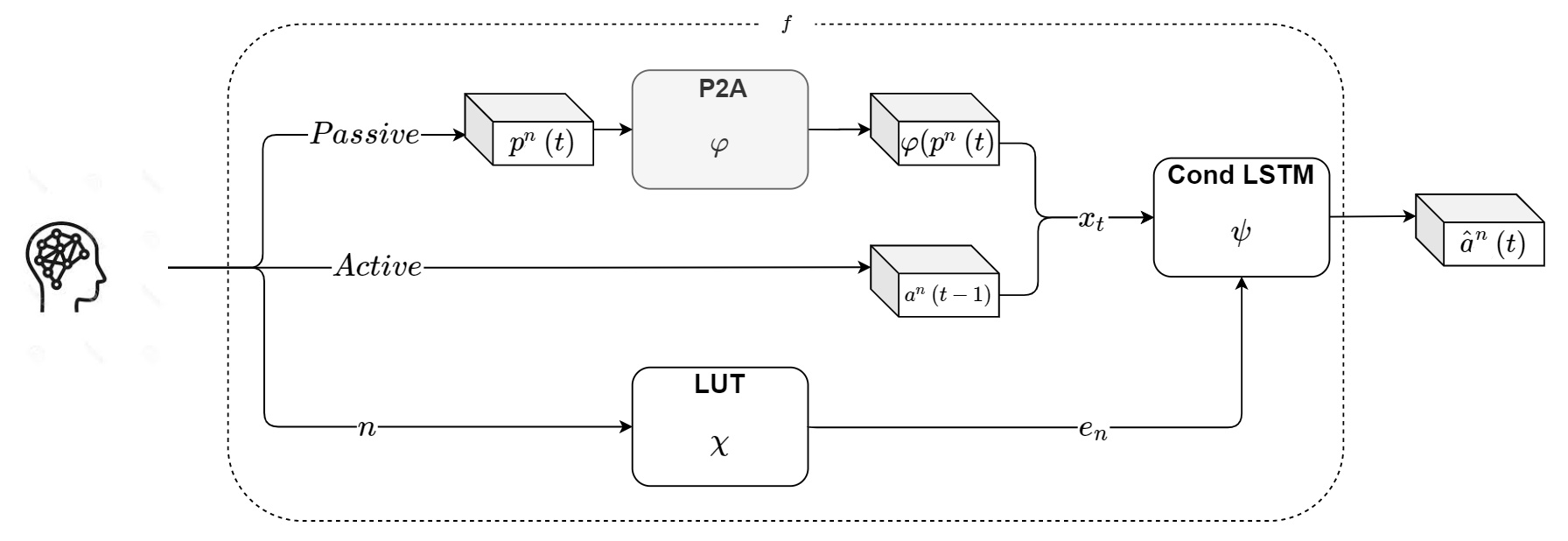}
    \caption{In our model's second phase of training, the Passive2Active translator, $\varphi$, is frozen. The conditional LSTM, $\psi$, and the LUT, $\chi$, are trained to predict $a^n(t)$, the current active frame, given \mbox{$x_t = (a^n[t-1], \varphi(p^n[t]))$}, and the subject identifier, $n$. The conditioning is based on a LUT, which provides the embedding vector of each subject, $\chi(n) = e_n$.}
    \label{fig: algo: active session prediction}
\end{figure}

When training the conditioned LSTM, the input, $x^n_t$, is a concatenation of $\big(\varphi(p^n[t]), a^n[t-1]\big)$. The model also receives $e_n=\chi(n)$, as a conditioning input. As illustrated in Fig.~\ref{fig: algo: active session prediction}, the subject's LUT, $\chi$, and conditioned LSTM, $\psi$, are tasked to predict the subject's current active frame, by minimizing the following loss function: 
\begin{equation}
    \mathcal{L}_{Recon} = \sum_n \sum_{t=0}^{M\cdot T}||a^n[t] - \psi(x^n_t, e_n)||_2
    \tag{11}
\end{equation}

In order to demonstrate that the user embedding is beneficial, we also train a baseline model, in which a vanilla LSTM is used, without the embedding.

\subsection{Predicting Psychiatric Traits using the subject's representation}

In the final training phase, after learning an embedding per subject, we freeze the LUT, and evaluate its utility in identifying psychiatric traits. For this purpose, we employ a linear classifier, denoted as $\rho$ (Eq.~\ref{eq:rho}), which is trained to minimize the Cross Entropy loss function, i.e, predicting the subject's psychiatric and demographic information, $y_n$, according to the subject's embedding, $e_n$.

The formulation as a classification problem and not as a regression problem is done in order to have the results of all prediction problems on the same human-interpretable scale.  The quantization of the three scores (TAS-20, STAI, CAPS-5) and age is done by calculating the mean and variance for every score, and creating five labels by the following ranges: $(-\infty, \mu-2\cdot\sigma]$, $(\mu-2\cdot\sigma, \mu-\sigma], (\mu-\sigma, \mu+\sigma]$, $(\mu+\sigma, \mu+2\sigma]$, $(\mu+2\sigma, \infty)$. Results for the regression scheme, presenting rMSE of the model trained with the L2 loss, are provided in the supplementary material.

\subsection{Inference}
When presenting our model with a new subject, $m$, we first use the already trained (and frozen) $f$ neural network to learn an embedding vector $e_m$. This is done using only the raw fMRI signals, by fitting $e_m$ in a next active frame prediction task. After $e_m$ is learned, it is passed through the linear classifier ($\rho$), to receive the model's prediction, $\hat{y}_m$.

\section{Experiments}
Data partitioning distinguishes between train, evaluation, and test set, each composed of different subjects, with a 60-20-20 split.
Each experiment was repeated 10 times on random splits, and our plots report mean and SD. In all of the below-mentioned experiments, the personal embedding vector size was set to 12 ($\mathbb{R}^{12}$).

\subsubsection{Next Active Frame Prediction}
\begin{figure}[t]
\begin{minipage}{0.6\textwidth}
    \centering
    \includegraphics[width=\textwidth]{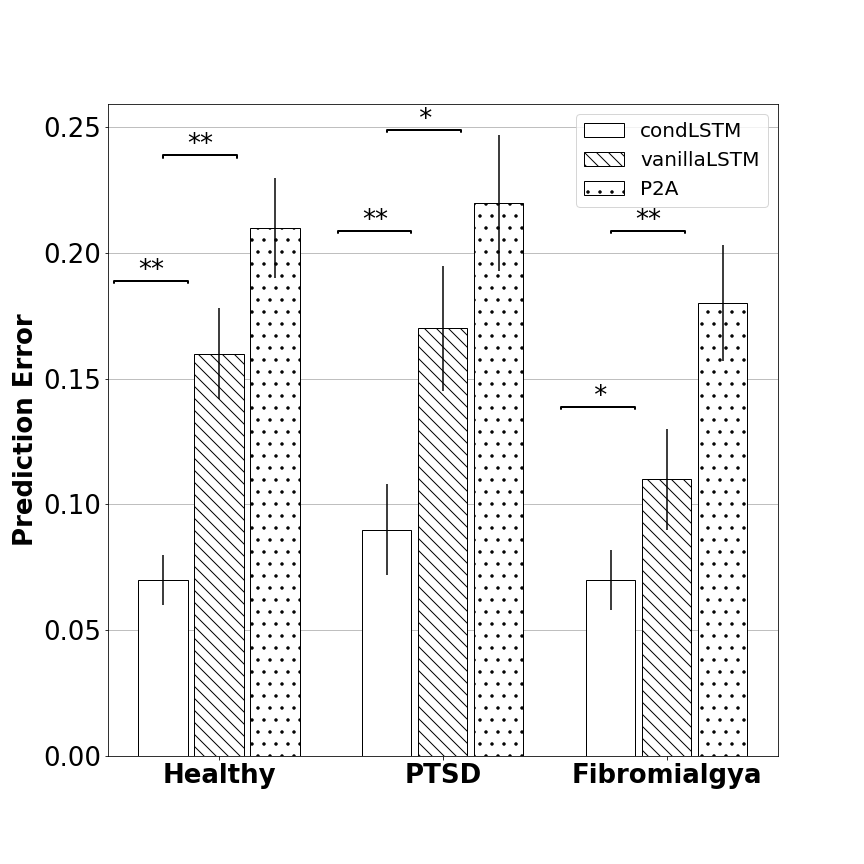}
    \caption{Next frame prediction error.\\ corrected re-sampled t-test~\cite{bouckaert2004evaluating}:\\ *$P < 0.05$, **$P<0.01$.}
    \label{fig:active frame error}
\end{minipage}
    \begin{minipage}{0.39\textwidth}
        \begin{minipage}{\textwidth}
            \includegraphics[width=1.2\textwidth]{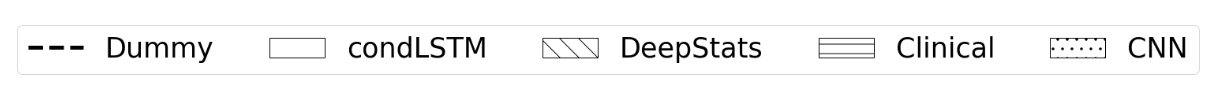}
        \end{minipage}
        \centering
        \begin{subfigure}{\textwidth}
            \includegraphics[width=\textwidth]{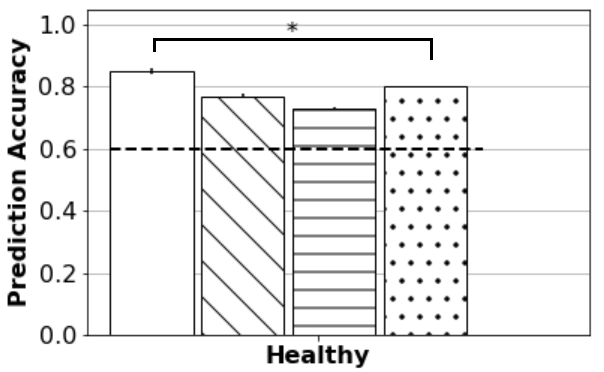}
            \caption{Past Experience}
        \end{subfigure}
        \begin{subfigure}{\textwidth}
            \includegraphics[width=\textwidth]{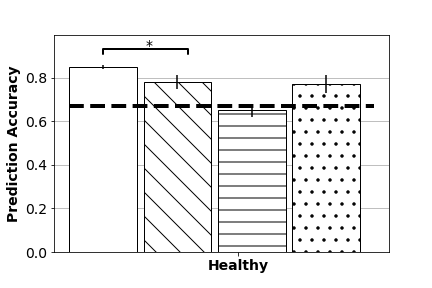}
            \caption{Age}
        \end{subfigure}
        \caption{Demographic accuracy}
        \label{fig: demography}
    \end{minipage}
\end{figure}

To evaluate the performance of our conditioned LSTM method, we compared {its performance in predicting the next active frame given the P2A output} to (i) a vanilla LSTM model with the same hidden-state size, and (ii) our trained Passive2Active Translator. Fig.~\ref{fig:active frame error} compares the three models. Evidently, the vanilla LSTM, which receives the data as a sequence of frames, significantly improves the performance of the memory-less Passive2Active Translator. Our conditioned LSTM, which incorporates the per-subject representation, outperforms the vanilla LSTM.

\subsubsection{Predicting Subject's Psychiatric and Demographic Criterias}
We test whether our learned representation, trained only with raw fMRI images, has the ability to predict a series of psychiatric and demographic criteria, not directly related to the neurofeedback task. 

We used our method to predict (i) STAI and (ii) TAS-20 for PTSD, Fibromiaglgya and control subjects, (iii) CAPS-5 for PTSD subjects. Demographic information, (iv) age, and (v) past NeuroFeedback experience were predicted for control subjects. 

Our linear classification scheme, applied to the learned embedding vectors, is compared to the following baselines, which all receive the raw fMRI sequence as input, denoted as $x$: (1) \textbf{fMRI CNN-} Convolutional layer with $k=10$ filters, followed by a mean pooling operation to receive $z(x)\in \mathbb{R}^{(2\cdot M, k, T)}$, which is the input to an MLP with $l=3$ layers, which predicts the label; (2) \textbf{fMRI Statistical Data}- This method performs two spacial pooling operations on the fMRI sequence: (i) mean and (ii) standard deviation to create the statistics tensor, $z(x) \in \mathbb{R}^{(2\cdot M, 2, T)}$, which is  fed to a linear layer, which predicts the label; (3) \textbf{Clinical Prediction-} SVM regression of every trait, according to the other traits (leave-one-trait-out, where the data contains all psychiatric traits and the two demographic traits); and (4) \textbf{Dummy Prediction-} Predicts the most common label.

\begin{figure}[t]
    \centering
    \begin{tabular}{ccc}
    \multicolumn{3}{c}{
        \includegraphics[trim=0 10 0 0,clip,width=0.8\textwidth]{Figures/legend.png}}\\
        \includegraphics[width=0.36\linewidth]{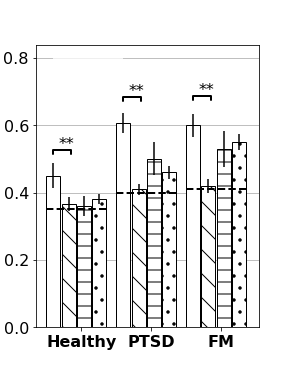}&
        \includegraphics[width=0.36\linewidth]{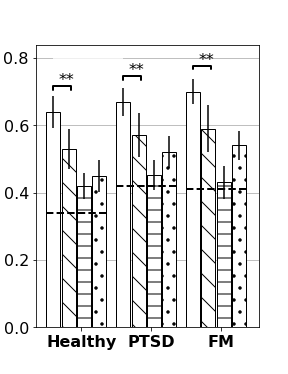}&
         \includegraphics[width=.23\linewidth]{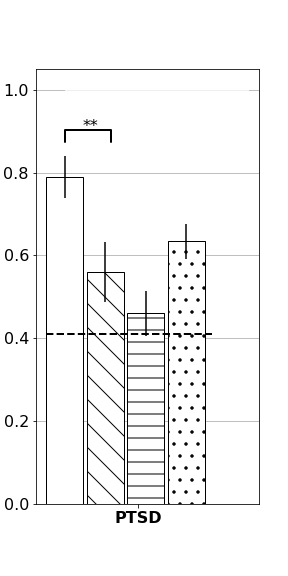}\\
    (a) & (b)&(c)\\
    \smallskip
        \end{tabular}

    \caption{Psychiatric trait predictions. (a) STAI, (b) TAS-20, (c) CAPS-5.}
    \label{fig: psyc}
\end{figure}

The results are shown in Fig.~\ref{fig: demography},\ref{fig: psyc}. It is evident that our method significantly outperforms the baseline methods in predicting the correct range of both the demographic and the psychiatric traits.

\section{Conclusions}
We present a method for learning a static, meaningful representation of a subject performing a neurofeedback task. This subject embedding is trained on a self supervised task and is shown to be highly predictive of psychiatric traits, for which no physical examinations or biological markers exist, as well as for age, and NF experience. We, therefore, open a new avenue for psychiatric diagnosis that is not based on an interview or a questionnaire.

\section*{Acknowledgments}
This project has received funding from the European Research Council (ERC) under the European Unions Horizon 2020 research and innovation programme (grant ERC CoG 725974). We thank Shira Reznik-Balter for insightful discussions on the analytic approach and helpful comments on the manuscript.

\bibliographystyle{splncs04}

\end{document}